\documentclass[nonacm, sigconf]{acmart}
\usepackage{makecell}
\AtBeginDocument{%
  }

\setcopyright{acmlicensed}
\copyrightyear{2026}
\acmYear{2026}
\acmDOI{XXXXXXX.XXXXXXX}
\renewcommand\footnotetextcopyrightpermission[1]{} % removes copyright footnote
\title[]{Your Title}
\acmSubmissionID{982}

\usepackage{multirow} 
\begin{document}

\title{MELCOT: A Hybrid Learning Architecture with Marginal Preservation for Matrix-Valued Regression }

%
% The "author" command and its associated commands are used to define
% the authors and their affiliations.
% Of note is the shared affiliation of the first two authors, and the
% "authornote" and "authornotemark" commands
% used to denote shared contribution to the research.
\author{Khang Tran}
\email{khangtranchicken2006@gmail.com}
\orcid{10009-0007-4753-0948}
\affiliation{%
  \institution{National University of Singapore}
  \country{Singapore}
}

\author{Hieu Cao}
\email{caothanhhieu2013@gmail.com}
\orcid{0009-0002-3618-529X}
\affiliation{%
  \institution{University of Science - VNUHCM}
  \city{Ho Chi Minh City}
  \country{Vietnam}}

\author{Thinh Pham}
\email{thinhcm2003@gmail.com}
\orcid{0009-0009-2465-6833}
\affiliation{%
  \institution{University of Science - VNUHCM}
  \city{Ho Chi Minh City}
  \country{Vietnam}
}

\author{Nghiem Diep}
\email{nghiemdt2013@gmail.com}
\orcid{0000-0001-7406-1250}
\affiliation{%
  \institution{University of Science - VNUHCM}
  \city{Ho Chi Minh City}
  \country{Vietnam}
}

\author{Tri Cao}
\authornote{Correspoding authors}
\email{tricao2001vn@gmail.com}
\orcid{0000-0001-7865-8476}
\affiliation{%
  \institution{National University of Singapore}
  \country{Singapore}
}

\author{Binh Nguyen}
\authornotemark[1]
\email{ngtbinh@hcmus.edu.vn}
\orcid{0000-0001-5249-9702}
\affiliation{%
  \institution{University of Science - VNUHCM}
  \city{Ho Chi Minh City}
  \country{Vietnam}
}

\renewcommand{\shortauthors}{Tran et al.}

\begin{abstract}

Regression is essential across many domains but remains challenging in high-dimensional settings, where existing methods often lose spatial structure or demand heavy storage. In this work, we address the problem of matrix-valued regression, where each sample is naturally represented as a matrix. We propose MELCOT, a hybrid model that integrates a classical machine learning–based Marginal Estimation (ME) block with a deep learning–based Learnable-Cost Optimal Transport (LCOT) block. The ME block estimates data marginals to preserve spatial information, while the LCOT block learns complex global features. This design enables MELCOT to inherit the strengths of both classical and deep learning methods. Extensive experiments across diverse datasets and domains demonstrate that MELCOT consistently outperforms all baselines while remaining highly efficient. 
\end{abstract}




\received{20 February 2007}
\received[revised]{12 March 2009}
\received[accepted]{5 June 2009}

\maketitle

\section{Introduction}
Regression is a core task in machine learning for modeling relationships between input and output variables. Matrix-valued regression (MVR) extends this paradigm by representing both inputs and outputs as matrices, thereby preserving row–column dependencies. This formulation naturally arises in many real-world applications. For example, in international sports analytics, one can model the relationship between country-level factors such as GDP, population, and life expectancy and the resulting medal distributions across different sports, naturally represented as input and output matrices. Despite its importance, a key challenge in MVR lies in designing models that can effectively preserve spatial structure while remaining computationally efficient.

Existing approaches to high-dimensional regression can be divided into three main families. Classical machine learning models such as SVM \cite{article}, random forests \cite{rigatti2017random}, and boosting methods are computationally efficient and interpretable but require vectorization, which inevitably discards important row–column interactions and spatial dependencies \cite{rong2018research, doan2015selecting}. Tensor-based methods, including CP, Tucker, and multilinear PLS \cite{faber2003recent, li2018tucker, zhang2017tensor}, together with their nonlinear or probabilistic extensions such as kernel ridge regression \cite{martin2001database}, Gaussian process regression \cite{zhao2015bayesian}, and random forest–based tensor regression \cite{huang2020robust}, better preserve multiway structure but remain largely linear or low-rank and therefore limited in expressiveness for complex data. Deep learning approaches, ranging from general neural architectures \cite{hussain2021regression, lathuiliere2019comprehensive} to CNN-based tensor regression networks and low-rank extensions \cite{kossaifi2020tensor, cao2017tensor}, currently define the state of the art, yet they are parameter-heavy, computationally expensive, and often difficult to scale effectively in real-world scenarios.

In light of these challenges, Optimal Transport (OT) provides a promising direction for regression with structured data. Originally introduced by Monge and later formalized by Kantorovich \citep{10.1112/plms/s1-14.1.139, 1370565168575910170}, OT offers a principled framework to align distributions while preserving structural information, and has been successfully applied in economics, imaging, and biology\cite{schiebinger2019optimal, rabin2015non, bousquet2017optimal}. Cuturi’s Sinkhorn algorithm \cite{cuturi2013sinkhorn} reduced the computational complexity of OT to practical levels, while recent advances that learn cost matrices directly from data \cite{stuart2020inverse, chiu2022discrete} have further enhanced its adaptability. These developments make OT particularly appealing for matrix-valued regression, where preserving local marginals while capturing global dependencies is essential.

In this work, we propose MELCOT, a hybrid architecture that integrates a machine learning–based Marginal Estimation (ME) block with a deep learning–based Learnable-Cost Optimal Transport (LCOT) block. The ME block preserves spatial information by estimating row and column marginals with lightweight computations, ensuring efficiency. The LCOT block leverages learnable OT to capture complex global dependencies, and can be parameterized with shallow neural networks to avoid excessive computational cost. This design allows MELCOT to balance efficiency with expressive power, overcoming the limitations of existing methods. Our contributions are three-fold: (1) We introduce MELCOT, a novel hybrid architecture that explicitly preserves marginal structure while modeling global dependencies through OT; (2) We conduct comprehensive experiments on the matrix-valued regression (MVR) problem, benchmarking MELCOT against classical, deep, and tensor-based baselines in both linear and nonlinear settings; and (3) We show that MELCOT consistently outperforms all baselines while maintaining high efficiency.  

\section{Preliminaries}\label{sec: pre}
\textbf{Optimal Transport (OT). }Created to provide a principled way to compare probability distributions by computing the most efficient plan to transform one distribution into another. Specifically, for a positive integer $n$, let $\mathbb{R}^n_+$ denote the space of $n$-dimensional vectors with non-negative real entries. Let $\mathbf{m}_1\in \mathbb{R}^{n_1}_+, \mathbf{m}_2 \in \mathbb{R}^{n_2}_+$ be two discrete probability distributions (i.e., vectors whose entries are non-negative and sum to 1). We define the feasible set $\mathcal{T}$ as follows:
\begin{align}
    \mathcal{T}(\mathbf{m}_1,\mathbf{m}_2)=\{\mathbf{X}\in\mathbb{R}^{n_1\times n_2}_+:\mathbf{X}\mathbf{1}_{n_2}=\mathbf{m}_1,\mathbf{X}^{\top}\mathbf{1}_{n_1}=\mathbf{m}_2\},
\end{align}
where $\mathbf{1}_n$ is a $n-$dimension vector with each entry equal to $1$.
Consider a cost matrix $\mathbf{C}\in\mathbb{R}_+^{n_1\times n_2}$, the Optimal Transport problem \cite{villani2008optimal} can now be formally expressed as finding:
\begin{align}\label{eq: main opt of OT}
    \text{OT}(\mathbf{m}_1,\mathbf{m}_2,\mathbf{C})=\min_{\mathbf{X}\in\mathcal{T}(\mathbf{m}_1,\mathbf{m}_2)}\langle \mathbf{C},\mathbf{X}\rangle ,
\end{align}
where $\langle \mathbf{A},\mathbf{B}\rangle=\text{Tr}(\mathbf{A}^{\top}\mathbf{B})$ is the Frobenius inner product. In this work, we employ the Sinkhorn algorithm \citep{cuturi2013sinkhorn} to compute the OT map. This solver incorporates an additional entropic regularization term, controlled by a parameter $\varepsilon$. Details on how this parameter is selected are provided in Section~\ref{sec: Exp}.\\
\textbf{Learnable-Cost Optimal Transport. }This method, also known as Inverse Optimal Transport \cite{stuart2020inverse,chiu2022discrete}, replaces a fixed, deterministic cost function with one that is learned directly from data through training. More formally, let $f_{\theta} : \Omega \rightarrow \mathbb{R}^{n_1 \times n_2}$ denote the \textit{cost function}, parameterized by $\theta$, where $\Omega$ is the data space. Let $\mathbf{X}_{\hat{\theta}}$ be the solution to Equation~\ref{eq: main opt of OT}, where the cost matrix $\mathbf{C}$ is computed via $f_{\hat{\theta}}$. The Learnable-Cost Optimal Transport problem aims to learn the optimal parameters $\theta$ by minimizing the discrepancy between the resulting transport plan and a ground-truth plan from training data:
\begin{figure*}[t]
    \centering
    \includegraphics[width=0.89\textwidth]{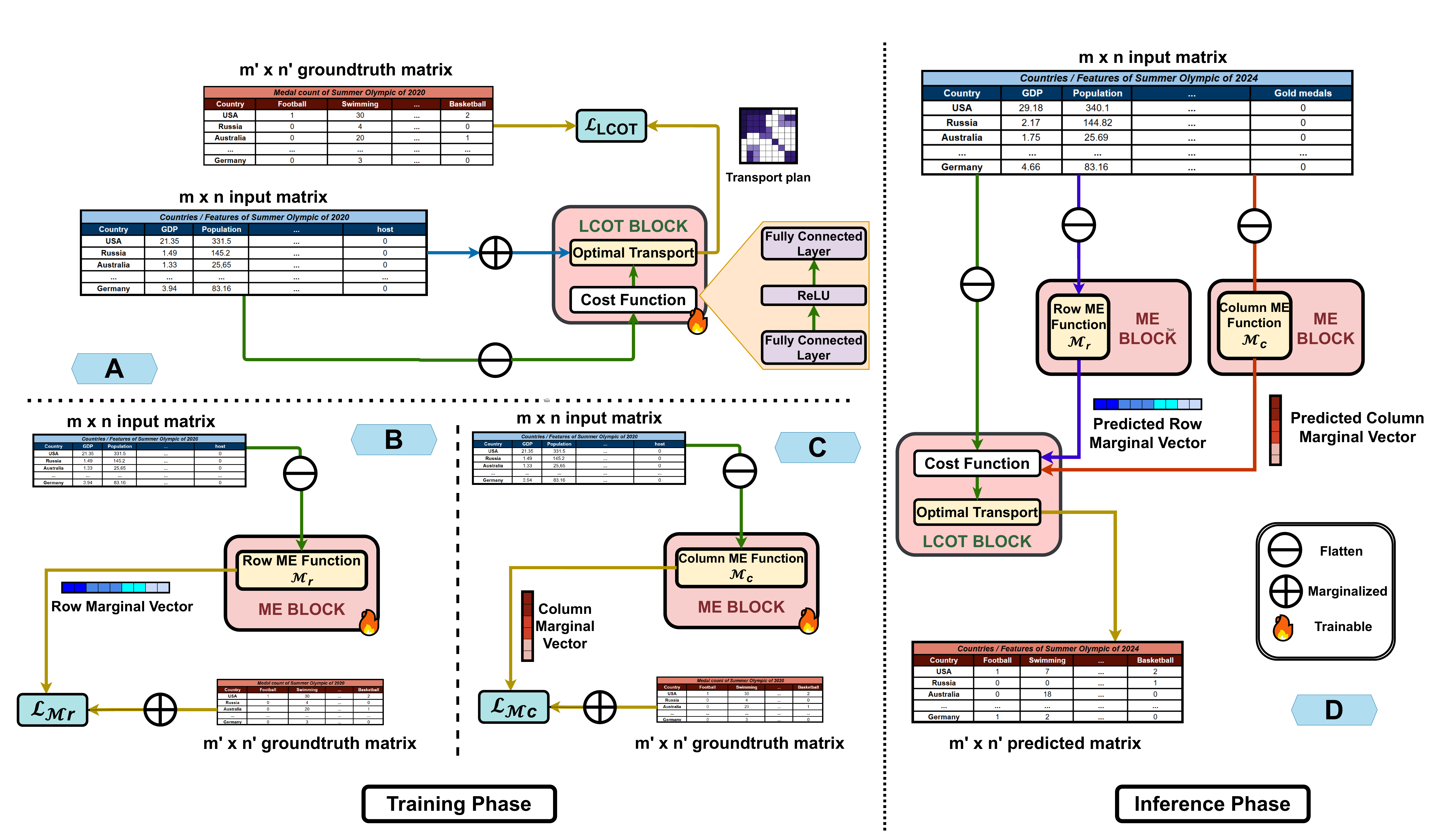} % adjust width as needed
    \caption{Overview of the MELCOT pipeline: (A) training phase of the LCOT block; (B) training phase of the ME row block $\mathcal{M}_r$; (C) training phase of the ME column block $\mathcal{M}_c$; and (D) inference phase of MELCOT. The pipeline begins with input data being used to train modules A, B, and C independently. During inference, blocks B and C are executed first to estimate the marginals, which, together with the original input data, are then passed into the trained LCOT block (A) to get the final prediction.}
    \label{Fig: Main archi}
\end{figure*}
\begin{align*}
    \arg\min_{\theta} \mathcal{L}(\mathbf{X}_{\theta}, \mathbf{X}_{\text{ground truth}}),
\end{align*}
where $\mathbf{X}_{\text{ground truth}}$ denotes the observed ground truth, and $\mathcal{L}$ is a predefined loss function measuring the alignment between the predicted value and the ground truth.\\
\textbf{Matrix-Valued Regression (MVR). }In the MVR problem, given a three-dimensional training dataset
$
  \mathcal{D}=\{(\mathbf{M}_t,\hat{\mathbf{M}}_t)\}_{t=1}^T,
$
where each sample \(\mathbf{M}_t\) is an \(m\times n\) matrix and \(\hat{\mathbf{M}}_t\) its corresponding \(m'\times n'\) label. Our goal is to learn a model function
$
  g:\mathbb{R}^{m\times n}\to\mathbb{R}^{m'\times n'}
$
that minimizes a pre-chosen loss \(\mathcal{L}\bigl(g(\mathbf{M}_t),\hat{\mathbf{M}}_t\bigr)\) over \(\mathcal{D}\). At test time, for a new input \(\mathbf{M}_{\mathrm{test}}\), we compute
\[
  \mathbf{M}_{\mathrm{pred}} = g(\mathbf{M}_{\mathrm{test}}),
\]
and evaluate performance of the model $g$ by comparing \(\mathbf{M}_{\mathrm{pred}}\) to the true label via different given metrics. Notably, in the special case \(n=1\), each \(\mathbf{M}_t\) collapses to a vector in \(\mathbb{R}^m\), recovering the standard regression setting. To address the MVR problem, conventional machine learning models typically flatten the data and reformulate it as a standard vector regression task. Specifically, each matrix‐valued pair $(\mathbf{M}_i,\hat{\mathbf{M}}_i)\in\mathbb{R}^{m\times n}\times\mathbb{R}^{m'\times n'}$ is flatten via $\mathbf{m}_i=\mathrm{vec}(\mathbf{M}_i)\in\mathbb{R}^{mn}$ and $\hat{\mathbf{m}}_i=\mathrm{vec}(\hat{\mathbf{M}}_i)\in\mathbb{R}^{m'n'}$.  The resulting dataset $\mathcal{D}'=\{(\mathbf{m}_t,\hat{\mathbf{m}}_t)\}_{t=1}^T$ is used to train the model $g:\mathbb{R}^{mn}\to\mathbb{R}^{m'n'}$ as a standard regression problem. The prediction result $\hat{\mathbf{m}}_{\mathrm{pred}}=g(\mathbf{m}_{\mathrm{test}})$  is then reshaped back to matrix form using the inverse mapping $\mathrm{mat}:\mathbb{R}^{m'n'}\to\mathbb{R}^{m'\times n'}$. However, this approach leads to the loss of spatial information as a result of the flattening and concatenation process. Hence, more advanced techniques, which we have introduced in the introduction, need to be developed to adapt to matrix data. 
\section{Main Architecture}\label{sec: main arch}
In this section, we describe the construction of our proposed architecture. We first present the design of each block, followed by the end-to-end training and testing procedure. The MELCOT framework comprises two main components: the ME block and the LCOT block, which are optimized independently during training and integrated at inference. An overview is provided in Figure~\ref{Fig: Main archi}.

\subsection{ME Block}\label{subsec: ME}
To ensure that our model preserves spatial information, we introduce the Marginal Estimation (ME) block. This block captures the row/column marginals of the target matrix so that spatial structure is retained and subsequently leveraged by the transport module in LCOT. In practice, some marginals may be known in advance (\textit{e.g.}, Olympic medal counts; see Section~\ref{sec: Exp}). In such cases, the ME block predicts only the unknown marginals, so that $\mathcal{M}_r$, $\mathcal{M}_c$, or neither requires training.\\ 
Formally, ME Block comprises two vector regressors: $\mathcal{M}_c$ for column margins and $\mathcal{M}_r$ for row margins. Given a training pair \(\mathbf{M}_i\in\mathbb{R}^{m\times n}\) and ground truth \(\widehat{\mathbf{M}}_i\in\mathbb{R}^{m'\times n'}\), we flatten \(\mathbf{M}_i\) into \(\mathbf{m}_i\in\mathbb{R}^{mn}\) and pass it through $\mathcal{M}_c$ and $\mathcal{M}_r$ to produce predicted marginals \(\mathbf{m}^c_{\text{pred}}\) and \(\mathbf{m}^r_{\text{pred}}\). Ground-truth marginals are obtained by marginalizing \(\widehat{\mathbf{M}}_i\), yielding \(\mathbf{m}^c_{\text{GT}}\) and \(\mathbf{m}^r_{\text{GT}}\). The ME losses are
\[
  \mathcal{L}_{\mathcal{M}_c}
  \bigl(\mathbf{m}^c_{\text{pred}},\,\mathbf{m}^c_{\text{GT}}\bigr), 
  \qquad 
  \mathcal{L}_{\mathcal{M}_r}
  \bigl(\mathbf{m}^r_{\text{pred}},\,\mathbf{m}^r_{\text{GT}}\bigr),
\]
and in our implementation, we use mean-squared error (MSE) for both $\mathcal{L}_{\mathcal{M}_c}$ and $\mathcal{L}_{\mathcal{M}_r}$.

\subsection{LCOT Block}\label{subsec: LCOT} 
The LCOT block reconstructs the full matrix by solving an optimal transport problem conditioned on the predicted marginals and a learned cost matrix. It consists of (i) a \emph{cost function} that maps the input to a cost matrix and (ii) an \emph{OT module} that solves the transport problem using these inputs.\\
\textbf{Cost Function.}
Having fixed the marginals (either predicted by the ME block or known a priori), the LCOT block learns a data-dependent transport cost. Concretely, for each training pair we take the same input matrix \(\mathbf{M}_i\in\mathbb{R}^{m\times n}\) and form its vectorization \(\mathbf{m}_i=\operatorname{vec}(\mathbf{M}_i)\in\mathbb{R}^{mn}\). This vector is fed to a two-layer multilayer perceptron, and the network output is reshaped to match the cost-matrix dimensions used by the OT module:
\[
  \mathbf{m}_i \xrightarrow{\text{2-layer MLP}} \operatorname{vec}(\mathbf{C}_{\text{pred}})\;\;\text{reshape}\;\;\rightarrow\;\;\mathbf{C}_{\text{pred}}\in\mathbb{R}^{m'\times n'}.
\]
Intuitively, \(\mathbf{m}_i\) summarizes the observed input structure, while the ME block supplies the target row/column totals; the cost network then maps \(\mathbf{m}_i\) to a matrix \(\mathbf{C}_{\text{pred}}\) that guides the OT plan consistent with those marginals. This cost function is the only learnable component in LCOT; it is trained jointly with the OT solver in the LCOT objective (see below), while the marginals used by OT come from ME (or prior knowledge) and are not backpropagated through this module.\\
\textbf{Optimal Transport Module.}
As in Subsection~\ref{sec: pre}, the module solves Eq.~\ref{eq: main opt of OT} given two marginals and the cost \(\mathbf{C}_{\text{pred}}\), returning a plan \(\mathbf{T}_i\in\mathbb{R}^{m'\times n'}\).
During training, \(\mathbf{m}_r,\mathbf{m}_c\) are the row/column sums of the ground truth \(\widehat{\mathbf{M}}_i\); at inference, they are either known a priori (e.g., Olympic counts) or predicted by the ME block.
The plan satisfies mass constraints \(\mathbf{T}_i\mathbf{1}_{n'}=\mathbf{m}_r,\;\mathbf{T}_i^\top\mathbf{1}_{m'}=\mathbf{m}_c\) and minimizes transport cost under \(\mathbf{C}_{\text{pred}}\).
For efficiency and differentiability, we use the Sinkhorn algorithm \cite{cuturi2013sinkhorn} with entropic regularization \(\varepsilon\) and tolerance \(\gamma\); ablations of OT solvers are in Table~\ref{tab: OT comp}.
Let \(\mathbf{T}_i\) denote the transport plan obtained by solving Equation~\ref{eq: main opt of OT} with inputs \((\mathbf{m}_r,\mathbf{m}_c,\mathbf{C}_{\text{pred}})\). The LCOT block is trained by minimizing
\[
  \mathcal{L}_{\mathrm{LCOT}}
  \bigl(\mathbf{T}_i,\,\widehat{\mathbf{M}}_i\bigr)
  \;=\; \|\mathbf{T}_i - \widehat{\mathbf{M}}_i\|_F.
\]

\subsection{Training and Testing Procedures}\label{subsec: TT}
\textbf{Training.}  
We optimize the ME and LCOT blocks independently. For each training pair \((\mathbf{M}_i,\widehat{\mathbf{M}}_i)\):
(i) ME: minimize \(\mathcal{L}_{\mathcal{M}_c}\) and \(\mathcal{L}_{\mathcal{M}_r}\) using inputs \(\mathbf{m}_i=\operatorname{vec}(\mathbf{M}_i)\) and targets from the marginals of \(\widehat{\mathbf{M}}_i\);
(ii) LCOT: predict \(\mathbf{C}_{\text{pred}}\) from \(\mathbf{m}_i\), solve Equation~\ref{eq: main opt of OT} with \((\mathbf{m}_r,\mathbf{m}_c,\mathbf{C}_{\text{pred}})\) (Sinkhorn with \(\varepsilon,\gamma\)), and minimize \(\mathcal{L}_{\mathrm{LCOT}}\).\\
\textbf{Testing.}  
Given a test sample \(\mathbf{M}\), we form \(\mathbf{m}=\operatorname{vec}(\mathbf{M})\). If prior marginals are available, we use them; otherwise, we obtain \(\mathbf{m}_{\mathrm{pred}}^r\) and \(\mathbf{m}_{\mathrm{pred}}^c\) from the ME block. In parallel, the cost function maps \(\mathbf{m}\) to \(\mathbf{C}_{\mathrm{pred}}\). The OT module then solves Equation~\ref{eq: main opt of OT} (via Sinkhorn) with these inputs to produce the transport plan \(\mathbf{T}_{\mathrm{pred}}\), which is compared to \(\widehat{\mathbf{M}}\) under the evaluation metrics. \\
For MELCOT to function, solving Equation~\ref{eq: main opt of OT} is essential in both training and testing; therefore, both the LCOT and ME blocks are compulsory components of the model. However, smaller elements within these blocks, such as the OT solver or the cost function parameterization, can be varied without compromising the overall framework.

\section{Experimental Results}\label{sec: Exp}

\begin{table}[t] 
  \centering
  \setlength{\tabcolsep}{1.6pt} 

   \scalebox{0.8}{
  \begin{tabular}{l|ccc|ccc|ccc}
    \toprule
    
    \multirow{2}{*}{\textbf{\huge Model}} & 
    \multicolumn{3}{c|}{\makecell{\textbf{Electricity}\\\textbf{Production}}} & 
    
    \multicolumn{3}{c|}{\textbf{Olympic Medal}} & \multicolumn{3}{c}{\textbf{Tourism}}\\
    \cmidrule(lr){2-4} \cmidrule(lr){5-7} \cmidrule(lr){8-10}
     & \textbf{rMSE}$\downarrow$ & \textbf{RSE}$\downarrow$ & \textbf{Time}$\downarrow$ & \textbf{rMSE}$\downarrow$ & \textbf{RSE}$\downarrow$& \textbf{Time}$\downarrow$ & \textbf{rMSE}$\downarrow$ & \textbf{RSE}$\downarrow$& \textbf{Time}$\downarrow$  \\

    \midrule
    XGBoost\cite{chen2015xgboost} & 0.009  & 0.25 & 27.2 & 0.60 & 0.52  & 183.8 & 0.100 & 1.27 & 11.3\\
    AdaBoost\cite{freund1997decision}  & 0.006 & 0.16 & 228.1 & 0.54 & 0.47 & 223.4 & 0.141  & 1.67 & 156.3\\
    RF \cite{rigatti2017random}& 0.007 & 0.19 & 32.7 & 0.57 & 0.49 & 225.9 & 0.141  & 1.60  & 12.9\\
    SVM \cite{article}& 0.010 & 0.32& 10.8 & 0.53  & \underline{0.46} & 86.2 & 0.100  & 1.38 & 5.2\\
    DT \cite{rokach2005decision} & 0.010 & 0.18 & 9.9 & 0.57 & 0.50 & 67.8 & 0.300  & 1.07 & 4.2\\
    \midrule
    DNN 1 layer & 0.024  & 0.65 & 1.5 & 1.06  & 0.92 & 1.7 & 0.632  & 7.46 & 1.4\\
    DNN 2 layer & 0.032 & 0.86 & 1.5 & 1.05  & 0.92  & 1.8 & 0.316 & 4.23  & 1.4\\
    TabNet \cite{arik2021tabnet} & 0.020 & 0.58  & 1.6 & 0.72 & 0.63 & 1.5 & 0.141 & 1.10  & 1.2 \\  
    FT-T \cite{gorishniy2021revisiting}& 0.020 & 0.45  & 43.1 & 0.54 & 0.48 & 28.7  & \underline{0.055}  & 0.66 & 1.6\\
    \midrule
    CP \cite{zhou2013tensor} & 0.013 & 0.36 & \underline{0.2} & 0.78 & 0.47 & \underline{0.5} & \underline{0.055} & \underline{0.62} & \underline{1.68}\\
    Tucker \cite{li2017parsimonious} & \underline{0.004} & \underline{0.12} & 1.3 & 0.77 & \underline{0.46} & 8.29 & 0.082 & 0.95 & 2.1\\
    PLS \cite{zhao2011multilinear} & 0.006 & 0.16 & \textbf{0.04}  & 0.83 & 0.50 & \textbf{0.08} &  0.069 & 0.80 & \textbf{0.06} \\
    \midrule 
    ResNetCP\cite{cao2017tensor} & 0.014 & 0.36 & 220.7 & 0.79 & 0.47 & 190.8 & 0.126 & 1.49 & 175.0 \\
    ResNetT\cite{cao2017tensor} & 0.020 & 0.54 & 326.9 & 0.77 & \underline{0.46} & 233.3 & 0.130 & 1.50 & 189.9 \\
    ResNetTT \cite{cao2017tensor} & 0.017 & 0.48 & 192.5 & 0.75 & \underline{0.46} & 184.4 & 0.133 & 1.53 & 173.4 \\
    \midrule
    \textit{MELCOT$_{SVM}$} & \underline{0.004} & \underline{0.12} & 5.9 & \underline{0.52} & \textbf{0.45} & 7.0 & \textbf{0.017} & \textbf{0.19} & 2.5 \\
    \textit{MELCOT$_{RF}$}  & \textbf{0.003}  & \textbf{0.05} & 16.7 & \textbf{0.51} & \textbf{0.45} & 17.7 & \textbf{0.017}  & \textbf{0.19} & 6.6\\
    \bottomrule
  \end{tabular}
  }
  \caption{Performance of MELCOT and baselines across datasets. MELCOT$_{SVM}$ and MELCOT$_{RF}$ denote MELCOT with the ME block implemented as SVM and RF, respectively. Models are evaluated on different metrics, including rMSE (root MSE), RSE, and inference time (ms). Best and second-best scores are highlighted in bold and underlined, respectively. }
  \label{tab: medal}
\end{table}

\begin{table}[t]
\scalebox{0.93}{
    \begin{tabular}{c|ccc}
\hline
\textbf{Method} & \makecell{\textbf{Training}\\\textbf{Time}}$\downarrow$ 
                & \makecell{\textbf{Inference}\\\textbf{Time}}$\downarrow$ 
                & \textbf{rMSE}$\downarrow$ 
                \\
\hline
\textbf{EOT}   & 0.03 & \textbf{0.002} & \textbf{0.017}  \\
\hline
\textbf{EPOT}  & 0.04 & 0.006          & 0.045            \\
\hline
\textbf{LPOT}  & \textbf{0.02} & 0.005 & 0.055 \\
\hline
\textbf{LPPOT} & \textbf{0.02} & 0.005          & 0.055            \\
\hline
\end{tabular}}

\caption{Ablation of different OT variants on the Tourism dataset within the OT module, evaluated by training time (seconds/iteration), inference time (seconds), and performance (rMSE). Best results are in bold.}
\label{tab: OT comp}
\end{table}

\textbf{Dataset.} We evaluate our approach on three datasets. The Olympic Medal dataset (compiled from \cite{worldbank_open_data} for economic indicators, \cite{ourworldindata_life_expectancy} for life expectancy, and \cite{comap2025mcm} for historical medal counts) predicts medal distributions across sports, with an input–output mapping of $\mathbb{R}^{13 \times 84} \rightarrow \mathbb{R}^{26 \times 84}$. Notably, in the Olympic Medal dataset, the column marginals are known in advance. These correspond to the total number of medals awarded in each sport, which remain constant across different years. The Electricity Production dataset \cite{tanwar_sustainable_energy} models energy production from fossil fuels, nuclear, and renewable sources for 104 countries, with shape $\mathbb{R}^{13 \times 104} \rightarrow \mathbb{R}^{3 \times 104}$. The Tourism dataset \cite{urban_tourism_impact} forecasts key tourism indicators (arrivals, departures, exports, receipts) across 31 countries, represented as tensors of $\mathbb{R}^{4 \times 31} \rightarrow \mathbb{R}^{4 \times 31}$.\\ 
\textbf{Baselines. }We compare MELCOT against four groups of baselines: (i) traditional machine learning models including XGBoost \cite{chen2015xgboost}, AdaBoost \cite{freund1997decision}, Random Forest (RF) \cite{rigatti2017random}, Support Vector Machine (SVM) \cite{article}, and Decision Tree (DT) \cite{rokach2005decision}; (ii) deep learning models including 1-layer and 2-layer fully connected networks, TabNet \cite{arik2021tabnet}, and Feature Tokenizer-Transformer (FT-T) \cite{gorishniy2021revisiting}; (iii) tensor-based models including Canonical Polyadic (CP) \cite{zhou2013tensor}, Tucker \cite{li2017parsimonious}, and Partial Least Squares (PLS) \cite{zhao2011multilinear}; and (iv) deep tensor-based models consisting of ResNet18 with a Tensor Regression Layer \cite{kossaifi2020tensor} and three low-rank tensor decomposition variants, CP, Tucker, and Tensor Train (TT) following \cite{cao2017tensor}, referred to in this paper as ResNetCP, ResNetT, and ResNetTT.\\
\textbf{Parameters and Fine-tuning.} For ME Block, we employ two approaches: SVM and Random Forest, each applied to both $\mathcal{M}_r$ and $\mathcal{M}_c$. For the LCOT block, the cost function is optimized with ADAM solver \cite{kingma2014adam} using a learning rate of $1 \times 10^{-2}$. The cost model is a 2-layer DNN with input and output dimensions of 10; the hidden dimension is dataset-specific: $30$ for the Electricity Production dataset and $10$ for the others. For OT module, we set the regularization coefficient to $\varepsilon = 5 \times 10^{-1}$, the convergence tolerance to $\gamma = 5 \times 10^{-5}$, and the maximum iterations to 1000.\\
% \textcolor{red}{Add RF} In all experiments, we use an SVM with regularization parameter $C = 2.7$ and epsilon-insensitive tube $\epsilon = 2$ for both $\mathcal{M}_r$ and $\mathcal{M}_c$ blocks.
\textbf{Results.} We use rMSE and RSE for evaluation metrics and compare inference time between baselines and our methods. The results, summarized in Table~\ref{tab: medal}, demonstrate that our proposed method consistently outperforms all baselines across the three datasets. On the \textit{Electricity Production} dataset, MELCOT achieves the lowest rMSE ($0.003$) and RSE ($0.05$), improving upon strong baselines such as PLS and Tucker. For the \textit{Olympic Medal} dataset, MELCOT again obtains the best results with rMSE of $0.51$ and RSE of $0.45$, outperforming tree-based models (AdaBoost) and tensor methods (CP). On the \textit{Tourism} dataset, MELCOT significantly surpasses baselines with an rMSE of $0.017$ and RSE of $0.19$, while most baselines yield $\text{RSE} > 1$. With respect to inference time, our method is only slower than small neural networks, whose simplicity makes them computationally lightweight, and traditional tensor methods, which reduce complexity by projecting data into low-dimensional representations. Nevertheless, our method consistently outperforms these approaches in terms of accuracy across the evaluated metrics.\\
\textbf{Ablation on OT variants. }In previous sections,  experiments were conducted using OT with the Sinkhorn solver. Nevertheless, OT admits several variants, among which is Partial Optimal Transport (POT). The key idea of POT is to transport only a portion of the total mass rather than the entirety. Formally, consider the same setting as in Subsection~\ref{sec: pre}, and define the feasible set of POT, denoted by $\mathcal{R}$, as follows:
\begin{align*}
    \mathcal{R}(\mathbf{m}_1,&\mathbf{m}_2,s)\\&=\{\mathbf{X}\in\mathbb{R}^{n_1\times n_2}_+:\mathbf{X}\mathbf{1}_{n_2}\leq\mathbf{m}_1,\mathbf{X}^{\top}\mathbf{1}_{n_1}\leq\mathbf{m}_2,\mathbf{1}_{n_1}^{\top}\mathbf{X}\mathbf{1}_{n_2}=s\},
\end{align*}
where $s$ is the transport mass. Consider a cost matrix $\mathbf{C}\in\mathbb{R}_+^{n_1\times n_2}$, the POT objective can now be written as:
\begin{align}\label{eq: POT LP}
    \text{POT}(\mathbf{m}_1,\mathbf{m}_2,\mathbf{C},s)=\min_{\mathbf{X}\in\mathcal{R}(\mathbf{m}_1,\mathbf{m}_2,s)}\langle \mathbf{C},\mathbf{X}\rangle ,
\end{align}
Furthermore, besides the Sinkhorn algorithm, both OT and POT can also be solved via Linear Programming (LP). For clarity, we denote Sinkhorn-based methods as EOT and EPOT, and LP-based methods as LPOT and LPPOT. We evaluate these four variants on the Tourism dataset using SVM for both $\mathcal{M}_r$ and $\mathcal{M}_c$, comparing training time, inference time, and performance (Table~\ref{tab: OT comp}). Overall, EOT, though slightly slower in training, achieves superior performance and inference efficiency compared to the other variants. \\
\textbf{Implementation Details. }All experiments are carried out on a system with an 8-core CPU and 32 GB RAM.

\section{Conclusion}\label{sec: concl}
In this work, we propose MELCOT, a hybrid architecture that combines a deep learning–based LCOT block with a machine learning–based ME block for matrix-valued regression. By estimating marginals independently and reconstructing the output via OT, MELCOT preserves spatial structure in the data. Experiments show that MELCOT consistently outperforms traditional deep learning, machine learning, and tensor-based methods across datasets of varying shapes, sizes, and noise levels. For future work, one can extend MELCOT to higher-dimensional settings or evaluate its performance in stochastic scenarios where data arrives sequentially.
\newpage
\section{GenAI Usage Disclosure}
In this work, we used ChatGPT and Grammarly to check for grammar errors and enhance clarity and coherence. While generative AI tools were employed during the writing process, the authors retain full responsibility for the content.

\bibliographystyle{ACM-Reference-Format}
\bibliography{sample-base}


\end{document}